\documentclass{article}
\usepackage{spconf,amsmath,graphicx}
\usepackage{booktabs}
\usepackage{caption}
\usepackage{subcaption}
\usepackage{framed,multirow}
\newcommand{\ra}[1]{\renewcommand{\arraystretch}{#1}}
\usepackage{url}

\usepackage{enumitem}
\setlist{nosep, leftmargin=14pt}
\usepackage{xspace}

\usepackage{color}
\definecolor{myred}{rgb}{.8,.0,.0}

\newcommand{\Amelia}{Amelia Jim\'{e}nez-S\'{a}nchez}

\newcommand{\chex}{CheXpert}
\newcommand{\cxr}{NIH-CXR14}
\def\eg{\textit{e.g.}~}

\def\etal{\textit{et al.}\xspace}
\def\ie{\textit{i.e.,}~}

\title{Detecting shortcuts in medical images - a case study in chest X-rays}

\name{\Amelia, Dovile Juodelyte, Bethany Chamberlain, Veronika Cheplygina}
\email{\texttt{\{amji,doju,bcha,vech\}@itu.dk}}

\address{Department of Computer Science, IT University of Copenhagen, Denmark}

\begin{document}
%
\maketitle

\begin{abstract}
The availability of large public datasets and the increased amount of computing power have shifted the interest of the medical community to high-performance algorithms. However, little attention is paid to the quality of the data and their annotations. High performance on benchmark datasets may be reported without considering possible shortcuts or artifacts in the data, besides, models are not tested on subpopulation groups. With this work, we aim to raise awareness about shortcuts problems. We validate previous findings, and present a case study on chest X-rays using two publicly available datasets. We share annotations for a subset of pneumothorax images with drains. We conclude with general recommendations for medical image classification. We make our code available\footnote{\scriptsize{\url{https://github.com/ameliajimenez/shortcuts-chest-xray}}}.
\end{abstract}
\begin{keywords}
Chest X-ray, pneumothorax, shortcuts, fairness, bias, validation, reproducibility
\end{keywords}
\section{Introduction}
\label{sec:intro}
\begin{figure}
\centering
\begin{tabular}{c@{\hspace{1.\tabcolsep}} c@{\hspace{1.\tabcolsep}} c}
    \includegraphics[width=0.3\linewidth]{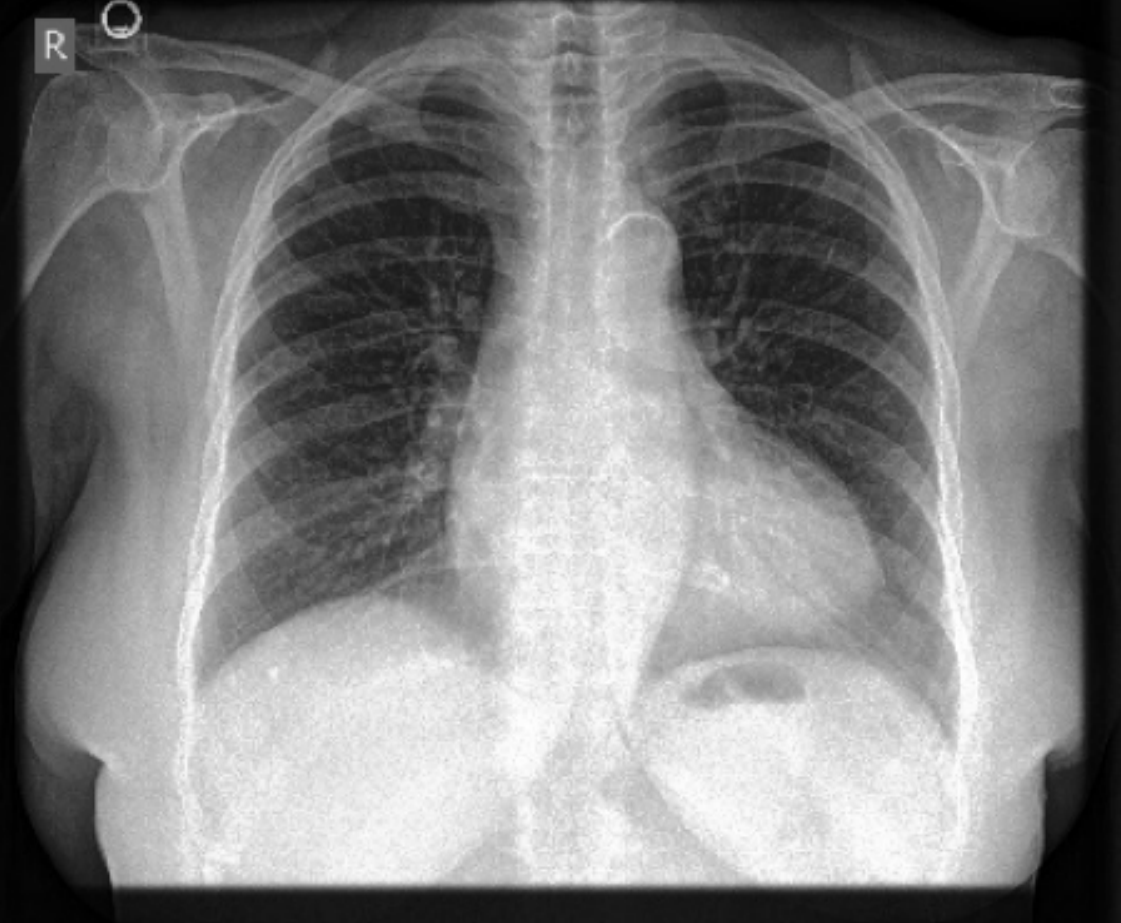}&
    \includegraphics[width=0.3\linewidth,height=2.12cm]{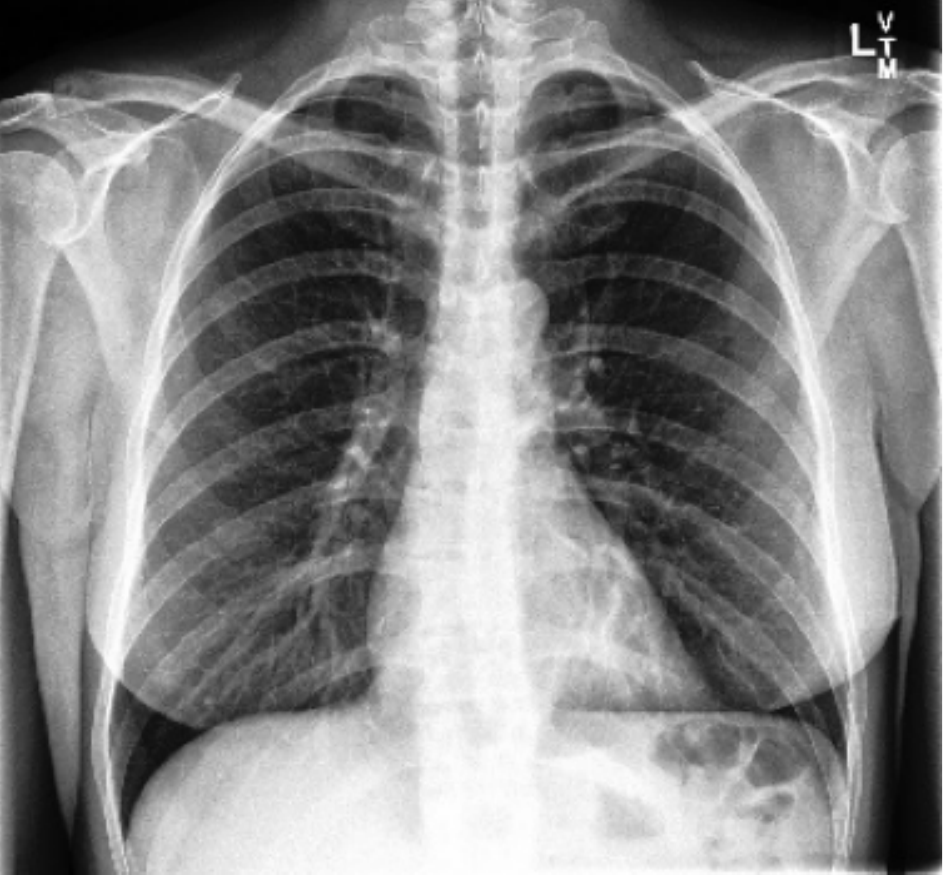}&
    \includegraphics[width=0.3\linewidth]{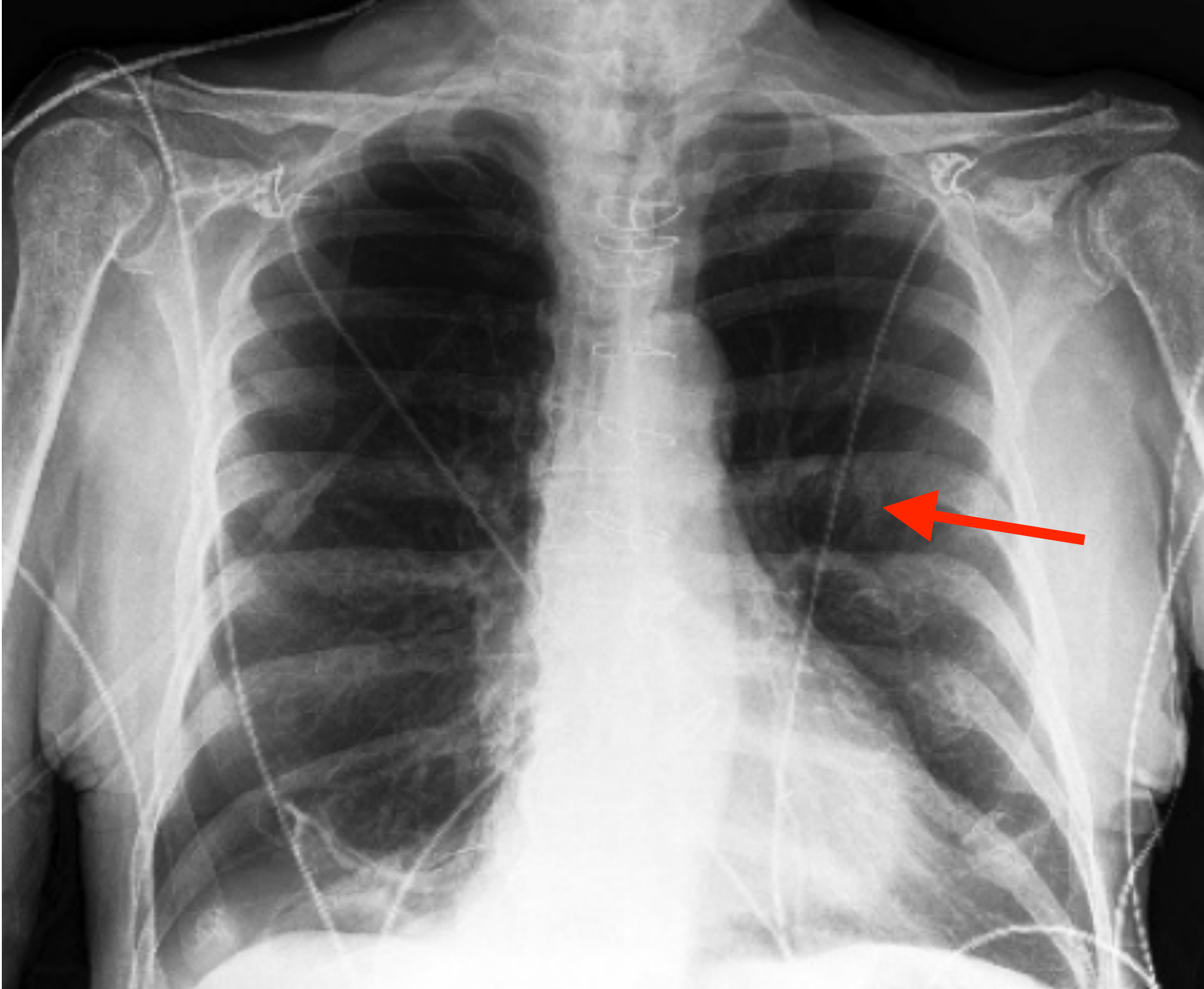}\\[2\tabcolsep]
    \includegraphics[width=0.3\linewidth]{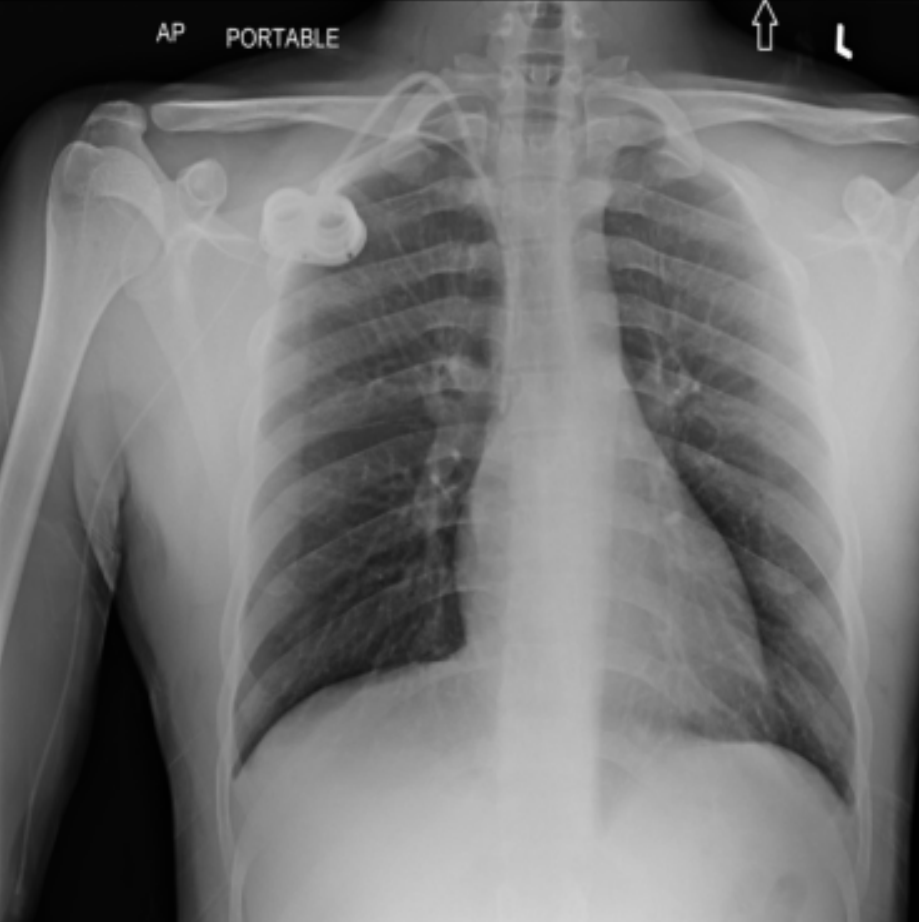}&
    \includegraphics[width=0.3\linewidth]{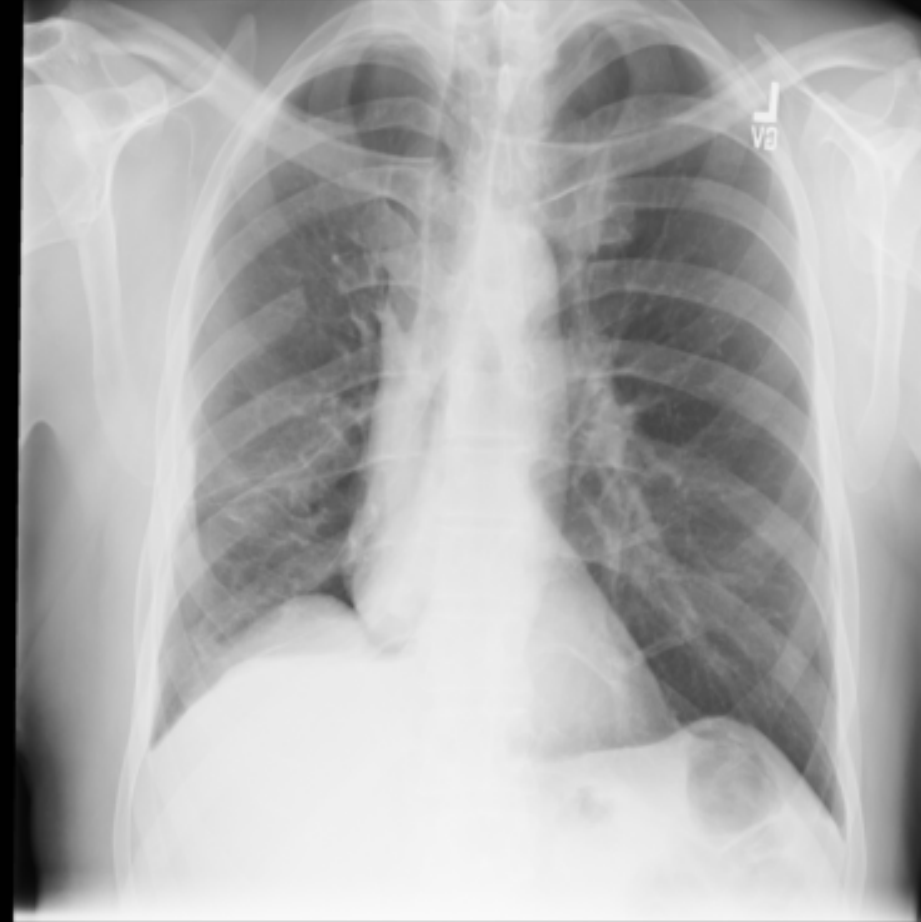}&
    \includegraphics[width=0.3\linewidth]{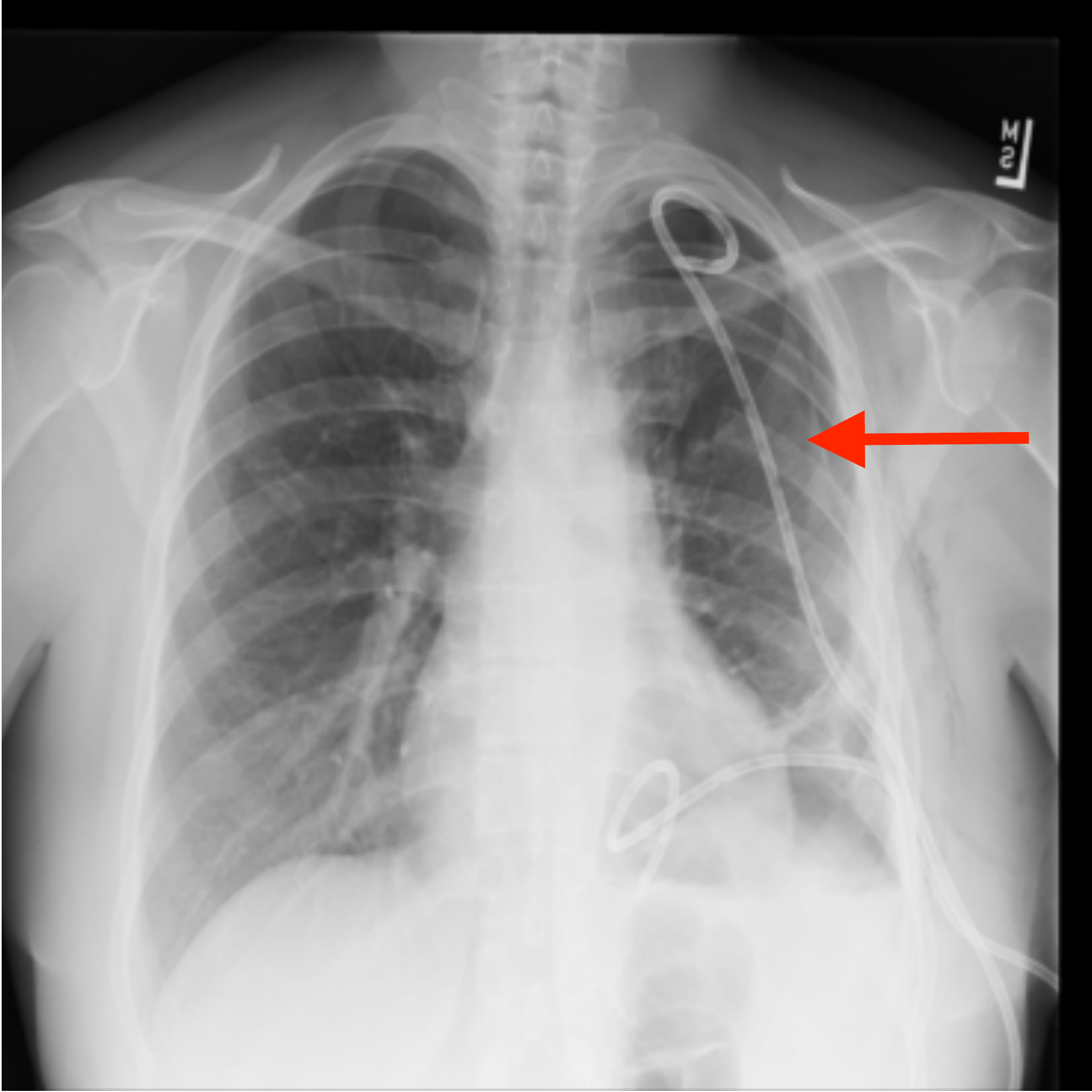}
\end{tabular}
\caption{Example of chest X-ray images from (left): negative pneumothorax, (middle): positive pneumothorax without drain, (right): positive pneumothorax with drain (red arrow). \\(Top): \chex{} and (bottom): \cxr{}.}
\label{figure:drains}
\end{figure}

Machine learning has shown promising results in medical image diagnosis, at times with claims of expert-level performance \cite{rajpurkar2017chexnet}. However, algorithms with high reported performances have been shown to suffer from overfitting on shortcuts, \ie spurious correlations between artifacts in images and diagnostic labels. Examples include pen marks in skin lesion classification \cite{winkler2019association}, patient position in detection of COVID-19 \cite{degrave2021ai}, and chest drains in pneumothorax (collapsed lung) classification \cite{oakden2020hidden}, see Fig.~\ref{figure:drains}. By training and evaluating on data with shortcuts, an algorithm's performance may appear high initially but will degrade when the shortcut is removed since it is unable to generalize based on relevant, diagnostic features.

Despite the current efforts, we find that there is not enough awareness of the issue overall, and thus, our motivation is to highlight the importance of detecting and mitigating shortcuts. For example, researchers may focus on high performance without realizing shortcuts might exist, while others may be aware but not have tools to reduce the impact of such shortcuts. The varied terminology (e.g. artifacts, shortcuts, bias, hidden stratification) further complicates finding related research.

Our contributions are as follows. First, we summarize methods to detect shortcut, by which we also collect the varied terminology for researchers to use when identifying related literature on this topic. Second, as an illustrative example of shortcuts, we present systematic experiments on \chex{} and \cxr{} that show degradation in performance when images with drains are excluded. This validates and generalizes (different data and methods) the findings of \cite{oakden2020hidden}. Third, as a byproduct of our experiments, we share a set of non-expert labels for chest drains, for a subset of \chex{} images with pneumothorax diagnosis. We conclude with general recommendations for medical image classification, and invite interested researchers to continue the conversation.

\begin{table*}[ht]  
\centering
\ra{1.3} 
\caption{(Left): distribution of image type for the three test scenarios. First symbol is for pneumothorax and second for the presence of a drain. Positive is represented by ``+" and negative by ``-". (Right): mean AUC (in \%) $\pm$ standard deviation for \chex{} and \cxr{}. Models are trained on development subsets (4k, 8k, 16k, 24k) of \chex{} dataset.}
\resizebox{17cm}{!}{
\begin{tabular}{@{}l|ccc|cccc|cccc@{}}
& \multicolumn{3}{c}{Image type} & \multicolumn{4}{c}{Train on \chex{}, test on \chex{}} & \multicolumn{4}{c}{Train on \chex{}, test on \cxr{}} \\
Scenario & +/+ & +/- & -/- & 4k & 8k & 16k & 24k & 4k & 8k & 16k & 24k \\ 
\midrule
baseline & 150 & 150 & 300 & 74.3 $\pm$ 0.3 & 77.1 $\pm$ 0.9 & 79.9 $\pm$ 0.6 & 81.0 $\pm$ 0.6 & 67.7 $\pm$ 3.1 & 70.8 $\pm$ 1.6 & 70.9 $\pm$ 3.8 & 74.4 $\pm$ 2.4 \\
w/o drain & 0 & 300 & 300 & 69.8 $\pm$ 0.9 & 72.2 $\pm$ 0.7 & 75.8 $\pm$ 0.9 & 76.7 $\pm$ 0.7 & 58.2 $\pm$ 2.5 & 63.3 $\pm$ 3.5 & 64.4 $\pm$ 3.8 & 65.5 $\pm$ 2.6\\
w/ drain & 300 & 0 & 300 & 80.2 $\pm$ 0.5 & 83.1 $\pm$ 1.8 & 84.8 $\pm$ 1.3 & 84.7 $\pm$ 0.8 & 75.3 $\pm$ 2.7 & 76.8 $\pm$ 2.4 & 76.8 $\pm$ 4.2 & 81.1 $\pm$ 3.5\\
\end{tabular}
}
\label{table:results}
\end{table*}

\section{Related Work}

\subsection{Detecting shortcuts}

The first step is to recognize that shortcuts might exist in the data, which can be done by inspecting the data itself, or studying the behavior of a trained model.

One work addressing both \textbf{data inspection} and \textbf{model inspection}, \cite{oakden2020hidden}, categorizes methods to detect shortcuts (here called hidden stratification) as \emph{schema completion}, \emph{error auditing} and \emph{algorithmic measurement}. Schema completion requires the labeling of a subset of the data. These type of methods are time consuming, subjective and limited by the author knowledge. Error auditing consists in observing the model outputs to find regularities. Error auditing approaches are also subjective, and critically dependent on the ability of the auditor to visually recognize differences in the distribution of model outputs. Algorithmic measurement relies on automatic subclass detection, for example with unsupervised methods such as clustering. These kind of methods still require human review, but are less dependent on the specific human auditor to initially identify the stratification. A considerable limitation of these approaches is on the separability of the important subsets in the feature space analyzed. 

Other recent works for detecting shortcuts focused on \textbf{model inspection}, and include skin lesion classification \cite{winkler2019association} or COVID-19 diagnosis \cite{degrave2021ai}. Winkler \etal{} \cite{winkler2019association} showed that adding skin markings to dermatoscopic images can cause a convolutional neural network (CNN) to flip its output. De Grave \etal{} \cite{degrave2021ai} found that CNNs learned the scanning position of COVID-19 patients as a shortcut for detection of the disease. They showed that the evaluation of a model on external data is not sufficient to ensure reliability, and concluded that more explainability is required to deploy systems in a clinical setting. \cite{Liu2022MedicalAlgoritmicAudit, Noseworthy2020AssessingMitigatingBias} propose several strategies for testing algorithmic errors, including exploratory error analysis, subgroup testing, and adversarial testing. A general recommendation is to report performance amongst diverse ethnic, racial, age and sex groups for all new systems to ensure a responsible use of machine learning in medicine.

\subsection{Reducing impact of shortcuts}
Once we are aware of the possible presence of shortcuts in our data, there are different bias mitigation methods we can employ to reduce their impact. Mehrabi~\etal~\cite{Mehrabi2021SurveyFairnessML} identify three main stages in which bias mitigation can be adopted: before (pre-processing), during (in-processing) and after (post-processing) training. 

\textbf{Pre-processing algorithms} mitigate bias in the training data. Strategies can consist of reweighting the training samples, or editing the features that will be used to increase fairness between the groups, \ie disparate impact remover \cite{Feldman2015DisparateImpactRemover}.

\textbf{In-processing algorithms} mitigate bias in the classifier. The target task loss is modified by adversarial learning, regularization or taking into account a fairness metric. Adversarial debiasing uses adversarial techniques to maximize accuracy and reduce the evidence of protected attributes in predictions~\cite{Zhang2018Mitigating}. The protected attributes are the characteristics for which fairness need to be ensured, such as gender or race. Prejudice remover adds a discrimination-aware regularization term to the learning objective \cite{Kamishima2012PrejudiceRemover}. Meta fair classifier takes the fairness as part of the input and returns a classifier optimized for the metric \cite{Celis2019MetaFair}.

\textbf{Post-processing algorithms} make predictions fairer. Reject option classification \cite{Kamiran2012RejectOption} modifies the predictions from the classifier, calibrated equalized odds \cite{Pleiss2017Calibrated} optimizes over the calibrated classifier's score outputs, and equalized odds modifies the prediction label using an optimization scheme.

\section{Methods}
 \subsection{Data} \label{subsec:data}
We use two publicly available chest X-ray datasets: \chex~\cite{Irvin2019Chexpert} and \cxr~\cite{Wang2017chestxray8}. 
\chex{} has 224,314 images from 65,240 patients and \cxr{} has  112,120 images from 30,805 patients. The sex distribution (male/female) is 59.4\%/40.6\% for \chex{} and  56.5\%/43.5\%, for \cxr{}. In both cases labels were extracted from radiology reports with natural language processing.

For \chex, two data science students labeled frontal images for the presence of chest drains. The students had no prior experience with medical imaging, and used \cite{jain2011pictorial,macduff2010management} to learn about the appearance of chest drains. They labeled the images independently, and reviewed the images where they disagreed. For the disagreement cases, they either reached agreement, or discarded the image. This process was continued until finding 300 pneumothorax positive images with chest drains. The students did not have access to the drain labels from \cxr. 

For \cxr, we use a subset of X-rays from the original test set. Labels were kindly provided by Lauren Oakden-Rayner, a board-certified radiologist and author of \cite{oakden2020hidden}.

\begin{figure*}
\centering
\begin{tabular}{c@{\hspace{1.\tabcolsep}} c@{\hspace{1.\tabcolsep}} c}
    \includegraphics[width=0.45\linewidth]{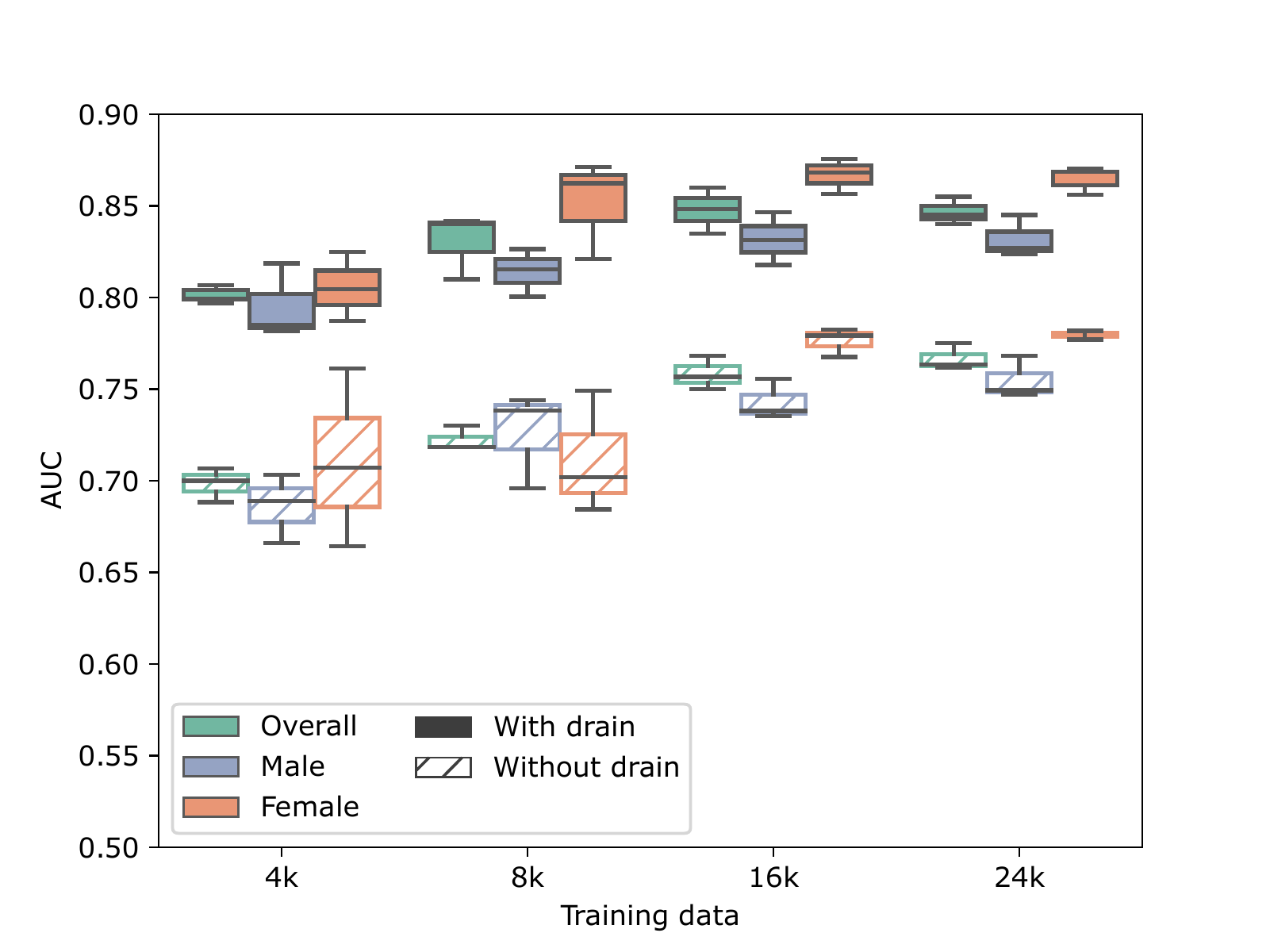} &
    \includegraphics[width=0.45\linewidth]{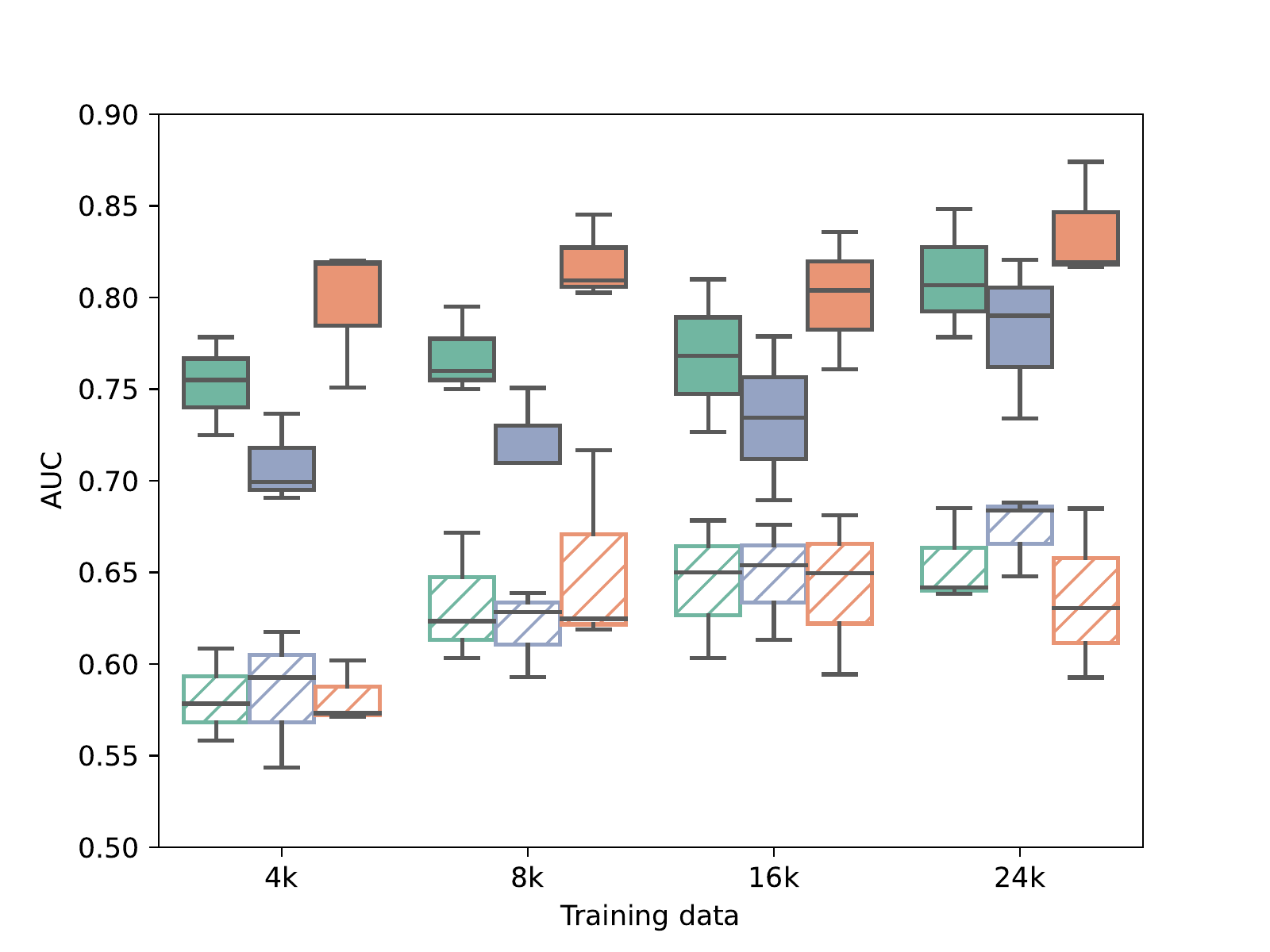}
\end{tabular}
\caption{Area Under the Curve (AUC) for the test scenarios: with drains and without drains. We show overall performance, as well as per sex attribute. (left): \chex{}, (right): \cxr{}.}
\label{figure:auc}
\end{figure*}

\subsection{Experiments} \label{subsec:experiments}
We focus on the pneumothorax classification task, following previous findings \cite{oakden2020hidden}. For the experiments, we train our models only with data from \chex. We train on differently sized development subsets (4k, 8k, 16k, 24k) with 50\% pneumothorax and 50\% other scans, resampling the training set 3 times. We adopt this strategy to account for variability in the training data and to get a distribution of performances for evaluation, as well as to avoid overly optimistic results that can be observed when fixed training/test splits are available~\cite{varoquaux2022machine}. Every development subset is split into 80\% training and 20\% validation.

We use the same model architecture as the original \chex{} model, namely a backbone CNN, followed by a probabilistic class activation map pooling and a fully connected layer. The backbone is a DenseNet-121, this CNN was found to be the best performing model in \cite{Irvin2019Chexpert}. We resize images to $512 \times 512$ px and train the models  for 10 epochs, with a batch size of 32, Adam optimizer and an initial learning rate of $1e-4$. Code used PyTorch~\cite{Paszke2019PyTorch} and Scikit-learn~\cite{Pedregosa2011Sklearn} libraries, and models were ran on an Nvidia v100 GPU at the ITU HPC cluster. We save the model with highest receiver-operating curve (AUC-ROC) on the validation set. 

We evaluate on both \chex{} and \cxr{}. We test on subsets of the relabeled images, varying the pneumothorax images from a mix between drains and no drains (baseline), with drains only (w/ drains), and without drains only (w/o drains), see Table \ref{table:results}. We report the overall AUC-ROC, and also for subgroups based on sex, following recent work on bias and fairness \cite{gichoya2022ai,larrazabal2020gender,seyyed2020chexclusion}. As additional evaluation, we use t-stochastic distributed neighbor embedding (t-SNE)~\cite{maaten2008visualizing} for understanding the representations learned by the trained networks. 


\section{Results}
We summarize our results in Table~\ref{table:results} and Fig.~\ref{figure:auc}. Firstly, we observe a decrease in the classification performance when testing on \cxr{}. This is similar to \cite{pooch2019can}, that reported an AUC of 0.87 for training and testing with \chex, and 0.74 for training on \chex{} and testing with \cxr.

For both datasets and the different development sets employed, we find that the classifier performs better on the subsets with drains. For example, training with 24k images, we obtain an AUC of 0.81 for the baseline, 0.85 for the subset with drains, and 0.77 for the subset without drains. This is in line with results on \cxr{} in \cite{oakden2020hidden}, with an AUC of 0.87 for pneumothorax, 0.94 with drains, and 0.77 without. 

Looking at the AUC differences by sex in Fig.~\ref{figure:auc}, we find overall higher AUCs for female patients. This suggests possible interactions that would need further investigation. Both \cite{larrazabal2020gender} and \cite{seyyed2020chexclusion} reported lower performances for female patients, but had different testing scenarios. 





\begin{figure}[t]
    \centering
    \includegraphics[width=0.8\linewidth]{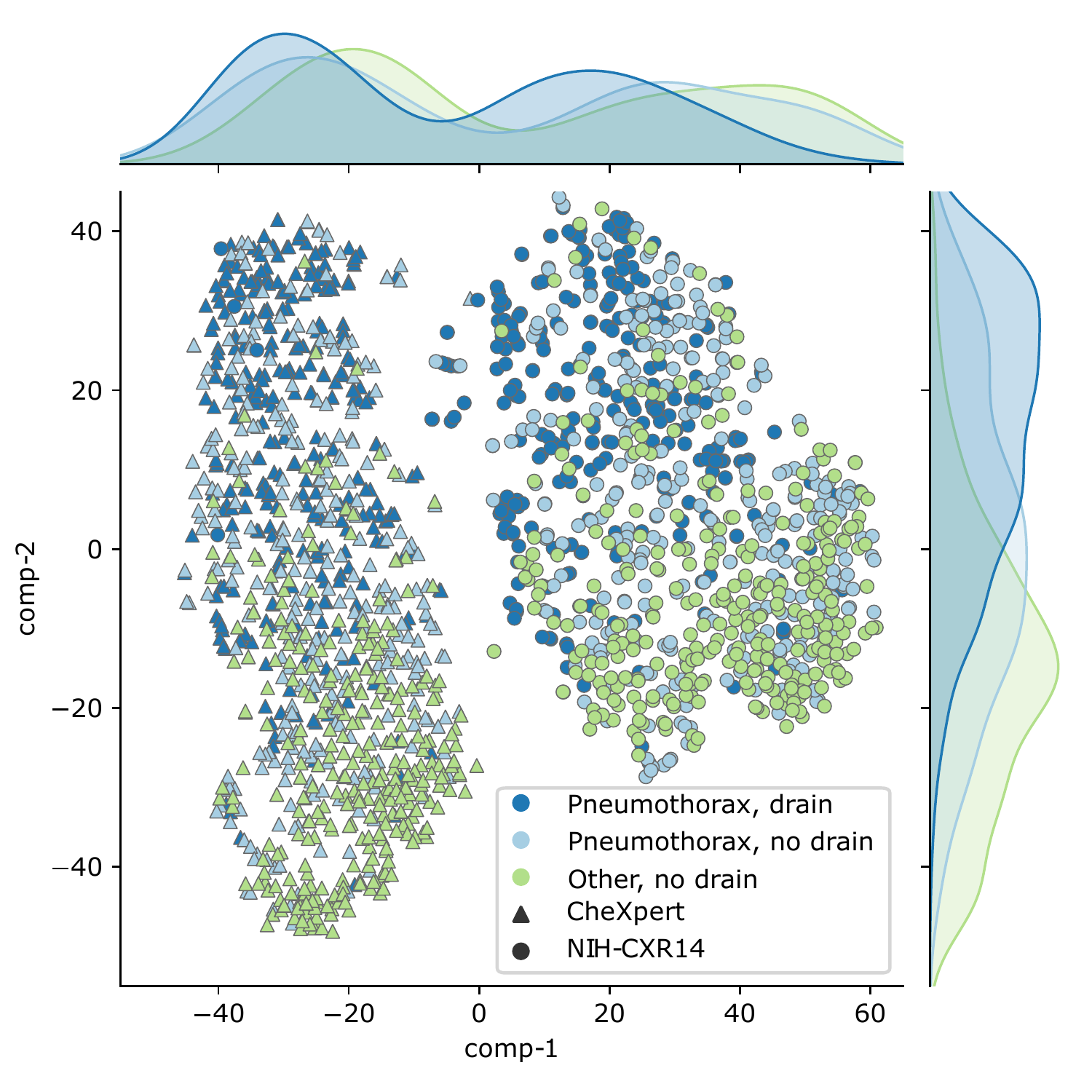}
    \caption{t-SNE projection of the feature embedding after the probabilistic class activation map of DenseNet-121.}
    \label{fig:tsne}
\end{figure}

Fig.~\ref{fig:tsne} displays the t-SNE projection of one of the models trained on the 24k development subset. From the embeddings, we see that in both datasets, there is a clear distinction between pneumothorax and other scans. The distribution of pneumothorax images with and without drains are more overlapping. There is a clear dataset shift between \chex{} and \cxr{}, with many positive/negative pneumothorax samples within a dataset closer to each other, than samples with the same label from the other dataset. Although the overall direction of class boundaries  appears to be the same, in practice even when training with more samples, the \chex{} boundary does not always fit the \cxr{} samples well, leading to larger variances in performance. 

\section{Discussion}

Our results validate earlier findings about CNNs memorizing chest drains as shortcuts for the pneumothorax label, despite using different training/test sets. We do not match earlier reported AUCs because we assess the variability in performance, rather than single estimates from earlier studies, which may be overly optimistic. 

We focused on pneumothorax classification due to findings in \cite{oakden2020hidden} and replicated the findings in a different dataset, with different (non-expert) annotators. We believe the problem is more general, and have preliminary results showing a similar degradation of performance in breast mammography classification, where the shortcut is text (the mammography view). This is an ongoing research direction, and we aim to expand this work to further applications.

We also plan to explore methods for adding or removing shortcuts, to be able to more systematically assess the extent of the problem in other applications. One could leverage powerful high-resolution image synthesis from text captions, such as Stable Diffusion (SD) \cite{Rombach2022StableDiffusion}. However, it is unclear what medical imaging concepts SD could incorporate, and potential ethical issues with memorization of private data may arise. 

A more general point for studies on shortcuts, is the chicken-and-egg relationship with large public datasets. Their size and availability allows for experimental comparisons, but is possibly also part of how shortcuts were introduced in the first place (for example, due to labels extracted by natural language processing). Since shortcuts might have interactions with demographic attributes, the medical imaging community should try to follow some general recommendations for fairness Artificial Intelligence \cite{Ganz2021AssessingBiasMedicalAI, Ricci2022AddressingFairness}. These recommendations point to a more extensive evaluation of the classifiers, for example by reporting metrics for relevant vulnerable subgroups such as the population stratified by \eg sex, age and ethnicity. 

In conclusion, shortcuts are an important problem in medical images that presents challenges to the robustness and fairness of algorithms. We invite others to continue this discussion, and welcome comments and suggestions for our future research.



\paragraph*{\textbf{Acknowledgments.}}
We thank Frederik Bechmann Faarup, Kasper Thorhauge Gr{\o}nbek, Andreas Skovdal (data science students) and Lauren Oakden-Rayner for early discussions and providing the labels. We thank Lottie Rosamund Greenwood for outstanding support on the ITU HPC cluster. This project has received funding from the Independent Research Fund Denmark - Inge Lehmann number 1134-00017B. 

\paragraph*{\textbf{Compliance with Ethical Standards.}} This research study was conducted retrospectively using human subject data made available in open access by the Stanford Hospital Institutional Review Board and NIH. Ethical approval was not required as confirmed by the license attached with the open access data.

\bibliographystyle{abbrv}
\bibliography{refs,refs_veronika}

\end{document}